%% file: main.tex
\documentclass[letterpaper, 10 pt, conference]{ieeeconf}  

\IEEEoverridecommandlockouts                              

\usepackage{amsmath} 
\usepackage{amssymb}  
\usepackage{xcolor}
\usepackage{algorithm}
\usepackage{algorithmic}
\usepackage{xspace}
\usepackage[small]{caption}
\usepackage{subcaption}
\usepackage{graphicx}
\usepackage{booktabs}
\usepackage{siunitx}
\usepackage{cite}

\newcommand{\VI}[1]{{\color{blue} #1}}

\newcommand{\guarantees}[1]{{\color{purple} #1}}
\newcommand{\black}[1]{{\color{black} #1}}
\newcommand{\brown}[1]{{\color{black} #1}}

\newcommand{\prob}[1]{\ensuremath{\mathbb{P}({#1})}}

\newtheorem{theorem}{Theorem}

\newtheorem{proposition}{Proposition}

\newtheorem{corollary}{Corollary}[theorem]

\title{\LARGE \bf
Data Association Aware POMDP Planning with Hypothesis Pruning Performance Guarantees}

\author{Moran Barenboim$^{1}$, Idan Lev-Yehudi$^{1}$ and Vadim Indelman$^{2}$
\thanks{$^{1}$Moran Barenboim and Idan Lev-Yehudi are with the Technion Autonomous Systems Program (TASP), Technion - Israel Institute of Technology, Haifa 32000,
	Israel, {\{\tt moranbar, idanlev \}@campus.technion.ac.il}}%
\thanks{$^{2}$Vadim Indelman is with the Department of Aerospace Engineering, Technion - Israel Institute of Technology, Haifa 32000, Israel. {\tt vadim.indelman@technion.ac.il}}
\thanks{This work was partially funded 	by US NSF/US-Israel BSF.} 
}

\begin{document}

\maketitle
\thispagestyle{empty}
\pagestyle{empty}

\input{00-Abstract.tex}


\section{INTRODUCTION}
\input{01-Introduction.tex} \label{sec:introduction}

\section{PRELIMINARIES}
\input{02-Preliminaries.tex}

\section{PLANNING WITH AMBIGUOUS DATA ASSOCIATIONS}
\input{03-Methods.tex}
\section{Mathematical analysis} \label{sec:analysis}
\input{04-Analysis.tex}

\section{EXPERIMENTS}\label{sec:experiments}
\input{05-Experiments.tex}

\section{CONCLUSIONS}
\input{06-Conclusions.tex}

\addtolength{\textheight}{-12cm}   








\bibliographystyle{IEEEtran}
\bibliography{refs}

\end{document}

%% file: 00-Abstract.tex
\begin{abstract}
    Autonomous agents that operate in the real world must often deal with
    partial observability, which is commonly modeled as partially observable
    Markov decision processes (POMDPs). However, traditional POMDP models rely
    on the assumption of complete knowledge of the observation source, known as
    fully observable data association. To address this limitation, we propose a
    planning algorithm that maintains multiple data association hypotheses,
    represented as a belief mixture, where each component corresponds to a
    different data association hypothesis. However, this method can lead to an
    exponential growth in the number of hypotheses, resulting in significant
    computational overhead. To overcome this challenge, we introduce a
    pruning-based approach for planning with ambiguous data associations. Our
    key contribution is to derive bounds between the value function based on the
    complete set of hypotheses and the value function based on a pruned-subset
    of the hypotheses, enabling us to establish a trade-off between
    computational efficiency and performance. We demonstrate how these bounds
    can both be used to certify any pruning heuristic in retrospect and
    propose a novel approach to determine which hypotheses to prune in order to
    ensure a predefined limit on the loss. We evaluate our approach in simulated
    environments and demonstrate its efficacy in handling multi-modal belief
    hypotheses with ambiguous data associations.
\end{abstract}


%% file: 01-Introduction.tex
Autonomous agents have become integral to our lives, from self-driving cars to
delivery robots. These agents must reason about partial observability when
interacting with the real world. For instance, an autonomous vehicle has to
reason about uncertain and incomplete information from its sensors to make
decisions such as choosing the correct lane or changing speed. Nevertheless,
most planning literature assumes complete knowledge of the source of the
observation, i.e., the observed environmental instance, but this may not be true
in practice. For example, self-driving cars use camera sensors to observe the
scene \brown{and relate surrounding objects to an a-priori known map. When a car
approaches a controlled intersection, it has to determine which of the visible
traffic lights correspond to the traffic light in the map and subsequently apply
to the lane it is driving. This is a simple problem if the localization is
perfect. However, sensor noise, changing lighting conditions, and occlusions can
cause the car to associate observations with an incorrect traffic light.}
Ignoring the possibility of inconsistent observation associations could lead to
an erroneous distribution shift of the state and potentially fatal consequences.

\begin{figure}[t]
	\begin{subfigure}{1\linewidth}
		\centering
		\includegraphics[scale=0.6]{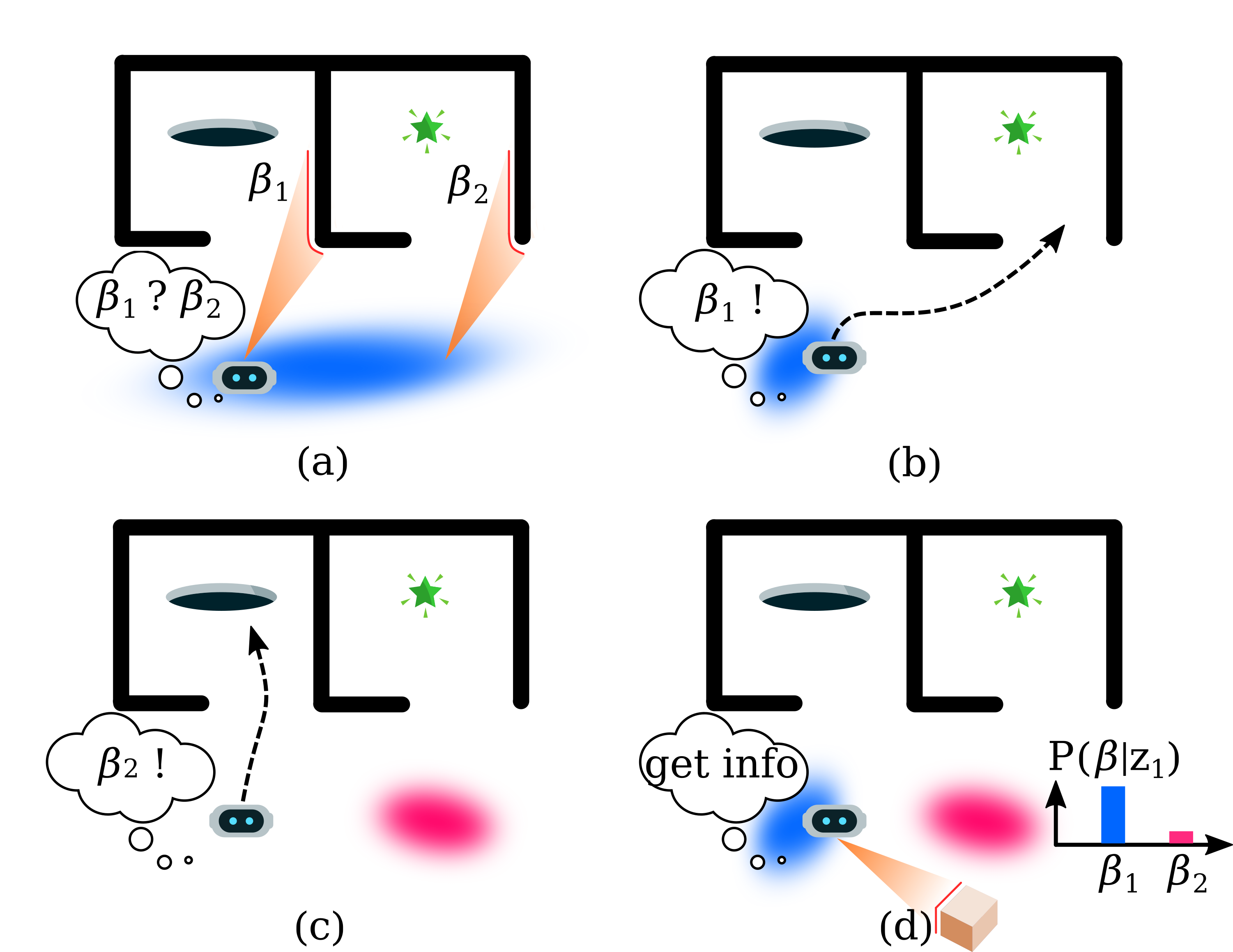}
	\end{subfigure}
	\caption{\scriptsize\brown{Figure (a) depicts an agent aiming to reach a
	goal (green star) while receiving an observation that could come from two
	sources, $\beta_1$ or $\beta_2$. In Figures (b) and (c), incorrect
	assumptions about the origin of the observation lead to changes in the
	robot's belief (blue and pink ellipses) and the optimal action, which can
	vary significantly. Notably, in (c), the calculated best action results in
	unsafe states. Instead, figure (d) showcases a data association aware belief
	and action, in which the agent holds two distinct hypotheses. Consequently,
	the agent chooses an action to gather information rather than traveling
	directly towards the goal.} \vspace{-12pt}}
	\label{fig:ambiguity}
\end{figure}

\brown{Figure \ref{fig:ambiguity} provides an example of a robot attempting to
reach a destination, represented as a star. In Figure \ref{fig:ambiguity}(a),
the robot perceives a potential future observation, but its exact pose is
unknown and expressed as a unimodal distribution. Equipped with a sensor having
a limited field of view, the robot detects a portion of a wall, which could be
part of a corridor leading to the goal (high reward) or a pit (low reward). In
Figures \ref{fig:ambiguity}(b) and (c), the robot assumes a deterministic source
for the observation, leading to potential selection of an incorrect and possibly
unsafe action. Figure \ref{fig:ambiguity}(d) demonstrates a multi-modal
posterior belief with different data association possibilities. Consequently,
the agent decides to gather more information rather than directly moving toward
the goal. This example highlights the importance of accounting for data
association ambiguity to avoid poor performance and unsafe policies where the
agent might mistakenly head towards the pit instead of the star.}

\brown{In general POMDPs, a plan that accounts for uncertainty maintains a
distribution over the possible states of the world. Accounting for ambiguous
data associations adds another layer of complexity by having to consider
multiple hypotheses, leading to a mixture distribution, where each component of
the mixture corresponds to a single hypothesis. Additionally, as the planning
horizon grows, the number of hypotheses grows exponentially \cite{Pathak18ijrr},
adding a significant computational burden.}

In response to the challenges posed by ambiguous data associations in POMDPs, we
propose a \brown{simplification approach, which maintains a small subset of the
hypotheses instead of maintaining an exponential number thereof.
Importantly, we derive bounds on the utility function between the POMDP with
the simplified and the non-simplified beliefs}. We use these bounds to establish
a trade-off between computational efficiency and performance for state-dependent
rewards. Further, using \brown{this relationship}, we propose a novel pruning
approach that balances computational efficiency with performance loss by
adaptively selecting which hypotheses to prune \brown{online}. 

Unlike current state-of-the-art POMDP planners that rely on particle
propagation, e.g. POMCP or DESPOT, our proposed approach overcomes the
challenge of particle depletion by introducing a novel estimator for the
objective function. This estimator is agnostic to the inference mechanism being
used, it supports \brown{both nonparametric and} parametric inference mechanisms
to enable long planning horizons. Through experiments in simulated environments,
we demonstrate the effectiveness of our proposed approach in handling
multi-modal belief hypotheses with ambiguous data associations.

In this paper we make the following main contributions: (a) we derive a theoretical
relation between the POMDP with a complete set of hypotheses and the pruned set
of hypotheses, enabling us to establish a trade-off between computational
efficiency and performance; (b) we develop an estimator that enables parametric
and nonparametric belief mixture representation to address particle depletion;
(c) we establish a similar relation between an estimated value function based on
the complete set of hypotheses and the value function of the pruned set of
hypotheses; (d) our bounds can be utilized to provide guarantees in terms of
worst-case loss in planning performance given some pruning method; (e) moreover,
we derive a scheme that utilizes our bounds to adaptively decide which
hypotheses to prune to meet a user-defined allowable loss in planning
performance. Finally, we demonstrate the effectiveness of our planning algorithm
in a simulated environment with unresolved data associations leading to
multi-modal belief. This paper is accompanied by supplementary
material \cite{Barenboim23ral_supplementary_b} that provides proofs for the
claims in this paper.

\section{\brown{Related Work}} While addressing the challenge of ambiguous data
associations (DA) has been extensively researched in the passive inference
community, \cite{Cadena16tro, Fourie16iros, Tchuiev19iros, Doherty19icra}, the
planning community has had relatively few attempts at supporting
\brown{ambiguous DA. General state-of-the-art POMDP planners, such as DESPOT,
POMCPOW or PFT-DPW \cite{Sunberg18icaps,Somani13nips} do not directly support DA
out-of-the-box. Although they can be altered to support DA, e.g. by replacing
the observation model with a mixture of observation models, an ad-hoc variation
will often result in particle depletion due to the multi-modal nature of a
multiple hypotheses belief. Particle depletion results in an overconfident and
potentially incorrect action selection due to the low representation of likely
state particles in a belief.}

\brown{A more dedicated approach for handling ambiguous DA could be to
explicitly maintain multiple representations of conditional beliefs, each
depending on different DA history. A naive attempt to perform planning with all
hypotheses results in an exponentially increasing number of hypotheses which is
computationally infeasible.} Instead, DA-BSP, \cite{Pathak18ijrr}, solves POMDPs
by explicitly maintaining hypotheses within the search tree and performs pruning
\brown{by keeping only a fixed number of the most promising hypotheses, or by keeping
only the hypotheses above some threshold on their probabilistic values.} However,
these pruning methods lack mathematical guarantees and are 
merely used as a tool to reduce the computational burden. More recently,
\cite{Shienman22icra, Shienman22isrr} considered different settings for planning
with hypotheses pruning and suggested an algorithm that actively plans to reduce
hypotheses ambiguity by defining an objective function over the hypotheses
distribution. Their approach provides bounds with respect to that unique
objective function and is specifically tailored for that task. Lastly,
\cite{Barenboim23ral_a} proposed an adaptive approach that
invests computational efforts in the most promising branches of both the
planning and hypotheses trees. Their method considers arbitrary state-dependent
rewards but comes only with asymptotic guarantees. 



%% file: 02-Preliminaries.tex
In this section, we formally define a POMDP with a belief that \brown{considers}
ambiguous data associations. The POMDP $M$ is a tuple
$\langle\mathcal{X}, \mathcal{A}, \mathcal{Z}, T, O, R\rangle$, where
$\mathcal{X}$, $\mathcal{A}$, and $\mathcal{Z}$ represent the state, action, and
observation spaces, respectively. The transition density function
$T(x_t,a_t,x_{t+1}) \triangleq \prob{x_{t+1} | x_t, a_t}$ defines the
probability of transitioning from state $x_t \in \mathcal{X}$ to state $x_{t+1}
\in \mathcal{X}$ by taking action $a_t \in \mathcal{A}$. The observation density
function $O(x_t,z_t) \triangleq \prob{z_t | x_t}$ expresses the probability of
receiving observation $z_t \in \mathcal{Z}$ from state $x_t \in \mathcal{X}$. 

Given the limited information provided by observations, the true state of the
agent is uncertain and a probability distribution function over the state space,
also known as a belief, is maintained. The belief depends on the entire history
of actions and observations, and is denoted $H_t \triangleq \{z_{1:t},
a_{0:t-1}\}$. We also define the propagated history as $H_t^- \triangleq
\{z_{1:t-1}, a_{0:t-1}\}$. At each time step $t$, the belief is updated using
Bayes' rule and the transition and observation models, given the previous action
$a_{t-1}$ and the current observation $z_t$, $b \left( x_t \right) = \eta_t
\int  
\prob{z_t | x_t } \prob{x_t | x_{t-1}, a_{t-1} } b \left( x_{t-1} \right) dx_t$,
where $\eta_t$ denotes a normalization constant and $b_t\triangleq\prob{x_t\mid
H_t}$ denotes the belief at time t. The updated belief, $b_t$ sometimes referred
to as the posterior belief, or simply the posterior. We will use them
interchangeably throughout the paper.

A policy function $a_t = \pi(b_t)$ determines the action to be taken at time
step $t$, based on the current belief $b_t$. In the rest of the paper we write
$\pi_t \equiv \pi(b_t)$ for conciseness. The reward is defined as an expectation
over a state-dependent function, $\rho(b_t, a_t)=\mathbb{E}_{x\sim b_t}[r_x(x,
a_t)]$. The value function for a policy $\pi$ over a finite horizon
$\mathcal{T}$ is defined as the expected cumulative reward received by executing
$\pi$,
\begin{equation} \label{eq: value function}
	V^{\pi }( b_{t}) =\rho( b_{t } ,\pi_t) + \underset{z_{t+1:T}}{\mathbb{E}} \left[ \sum _{\tau =t+1}^{T}\rho( b_{\tau } ,\pi _\tau)\right].
\end{equation}
The action-value function is defined by executing action $a_t$
and then following policy $\pi$ for a finite horizon $T$. The goal of
the agent is to find the optimal policy $\pi^*$ that maximizes the value
function.

\subsection{Ambiguous Data Associations as Mixture Belief}
To represent ambiguous data associations within the POMDP framework we define
the belief as a mixture distribution, that encompasses both continuous and
discrete random variables. The discrete variables, $\beta_{t}$, represent
different associations to seen observations at time $t$.  We formally define the
mixture belief at each time $t$ as,
\begin{equation} \label{eq: mixture belief}
    b \left( x_t \right) = \sum_{\beta_{0:t}}\prob{\beta_{0:t} \mid H_t}\prob{x_t | \beta_{0:t}, H_t},
\end{equation}
where $\prob{\beta_{0:t} \mid H_t}$ is the marginal belief over discrete
variables which can be considered as the mixture weight. An hypothesis,
$\beta_{0:t}$, denote the entire sequence of associations up to time step $t$.
$\prob{x_t | \beta_{0:t}, H_t}$ is the
conditional belief over continuous variables, given that the history and
associations are known. The marginal belief over the hypothesis, $\beta_{0:t}$,
can be updated by applying Bayes rule followed by chain rule,
\begin{align} \label{eq: recursive weight update}
&\prob{\beta_{0:t} \mid H_t} = \eta_t \mathbb{P}(z_t \mid \beta_{0:t}, H_t^-) \mathbb{P}(\beta_{0:t}\mid H_t^-)\\
&= \eta_t \mathbb{P}(z_t \mid \beta_{0:t}, H_t^-) \mathbb{P}(\beta_t \mid \beta_{0:t-1}, H_t^-) \mathbb{P}(\beta_{0:t-1}\mid H_t^-). \notag
\end{align} \normalsize
The conditional belief is updated for each realization of discrete random
variables as
\begin{equation}\label{eq: conBeliefUp}
    \prob{x_t | \beta_{0:t}, H_t} = \psi\big(\prob{x_{t-1} | \beta_{0:t-1}, H_{t-1}}, a_{t-1}, z_t\big),
\end{equation}
where $\psi(.)$ represents the Bayesian inference method. Last, the reward
function can now be written in terms of hypothesis dependency, $r(b_t,
a_t)=\mathbb{E}_{x\sim b_t}[r_x(x,
a_t)]=\mathbb{E}_{\beta_{0:t}}\left[\mathbb{E}_{x}[r_x(x, a_t)\mid
\beta_{0:t}]\right]$. For conciseness, we will denote 
\begin{equation} \label{eq:reward_beta}
    r(b_t^\beta,\pi_t) \triangleq \mathbb{E}_{x}[r_x(x, a_t)\mid
\beta_{0:t}].
\end{equation}

\subsection{IS and SN estimators}
Importance sampling (IS) is a Monte Carlo simulation technique for estimating
the expected value of a target function with respect to a probability
distribution. The IS estimator involves drawing samples from a proposed
distribution and weighting them by the ratio of the target distribution,
$\prob{\cdot}$ to the proposal distribution, $Q(\cdot)$,
\begin{equation}
    \hat{\mathbb{E}}^{IS}\left[r(x)\right]\triangleq\frac{1}{N}\sum_{i=1}^N \omega(x^i) r(x^i)=\frac{1}{N}\sum_{i=1}^N \frac{\prob{x^i}}{Q(x^i)} r(x^i).
\end{equation}
The estimator is unbiased and consistent \cite{Doucet00}, when the proposal
distribution is non-zero wherever the target distribution is non-zero.
Self-normalized importance sampling sometimes serves as a lower-variance
estimator by normalizing the importance weights. The SN-estimator is described
as,
\begin{equation}
    \hat{\mathbb{E}}^{SN}\left[r(x)\right] \triangleq \sum_{i=1}^N \frac{\omega(x^i)}{\sum_{j=1}^N\omega(x^j)} r(x^i),
\end{equation}
which converts the weights to a probability distribution. The SN-estimator is
biased, but consistent estimator.

%% file: 03-Methods.tex
In this section, we provide an overview of our algorithm, DA-MCTS, and the
baseline algorithm, vanilla Hybrid Belief-MCTS (HB-MCTS) \cite{Barenboim23ral_a}.
To facilitate understanding, we present the pseudo-code for both algorithms
jointly in Algorithm \ref{alg:algorithm}. We adopt a unified view, with
comments indicating the lines unique to each algorithm.

DA-MCTS is built upon the vanilla HB-MCTS algorithm, which itself is an
adaptation of PFT-DPW \cite{Sunberg18icaps} and MCTS \cite{Kocsis06ecml}. While
we have chosen to use these algorithms as the foundation for our work, we
acknowledge that other approaches may also be applicable, and we leave
exploration of these avenues to future research.

Vanilla HB-MCTS, a variant of belief-Markov Decision Process (BMDP), reframes
the POMDP into a belief-state model. In this, states are replaced by
belief-states reflecting an agent's environmental uncertainty. The transition
and observation functions update prior to posterior beliefs based on action and
observation, mirroring the stochastic state changes in a standard MDP. By
transforming POMDP to a BMDP, many MDP planning algorithms, including MCTS, can
be used as planning solvers. Notably, single particle propagation algorithms,
such as POMCPOW, are also possible, but may suffer from particle depletion as
mentioned in section \ref{sec:introduction}.

Algorithm \ref{alg:algorithm} presents a pseudo-code for the vanilla HB-MCTS
algorithm. In the \textsc{Simulate} procedure, an action is selected based on
the Upper Confidence Bound (UCB) heuristic in line \ref{line:UCB}. Depending on
whether the budget on the number of observations has been met, the algorithm
either expands a new posterior node, which includes its belief and reward
function, and then performs a rollout, or uniformly samples an existing
posterior node and continues recursively to the next node. Finally, the action
value of the current node and its relevant counters are updated. The vanilla
HB-MCTS algorithm is flexible in that the number of maintained posterior
hypotheses can be controlled and remain fixed based on a pre-defined
hyperparameter. For instance, a vanilla HB-MCTS with low compute resources can
have a pruning budget, where only $K$ hypotheses are maintained in each node of
the planning tree. The pruned hypotheses are usually chosen heuristically, e.g.
based on their probability value.

However, Vanilla HB-MCTS is limited in its ability to provide guarantees when
pruning is performed. While the performance guarantees we present in the next
section are applicable to any pruning heuristic, such as the one used in vanilla
HB-MCTS, we introduce a slightly different approach. Instead of pre-defining a
fixed number of hypotheses to maintain, we propose an adaptive approach that
determines which hypotheses to prune online based on a pre-defined maximum
allowable loss, $ \epsilon_{\bar{D}}$. We then modify the HB-MCTS algorithm to
adaptively determine which hypotheses to prune, while maintaining performance
guarantees with respect to the complete set of hypotheses. This modification is
reflected in line \ref{line:apriori}.

In addition, DA-MCTS can provide even tighter guarantees in hindsight without
incurring additional computational complexity, denoted by
$\hat{\epsilon}^{hs}_{\bar{D}}$, shown in line \ref{line:posteriori}. The increased
accuracy of these guarantees is due to the granularity of the hypotheses
weights. For instance, when there is only a single hypothesis, no hypotheses are
pruned, resulting in zero additional loss to the value function. The specific
bounds and estimators used are discussed in the following section. 

\begin{algorithm}[ht]
    {\scriptsize
        \caption{\VI{HB-MCTS} and \guarantees{DA-MCTS}}
        \label{alg:algorithm}
        \textbf{Procedure}:\textsc{Simulate}($b,h,d,\epsilon_{\bar{D}}$)\\
        \black{/*Init: $N(b), N(ba), Q(ba), \hat{\epsilon}_{\bar{D}}^{hs}(b),
        \hat{\delta}_{\bar{D}}^{\beta}(b)$ to $0$*/}
        \begin{algorithmic}[1] 
            \IF{d = 0}
            \RETURN 0
            \ENDIF
            \STATE $a \xleftarrow{} \underset{\bar{a}}{\arg \max} \ Q(b\bar{a}) +
            c\sqrt{\frac{log(N(b))}{N(b\bar{a})}}$ \label{line:UCB}
            \IF{$|C(ba)| \leq k_oN(ba)^{\alpha_o}$ }
            \STATE \VI{$b' \xleftarrow{}$ \textsc{PrunedPosterior}$(b, a)$
            /*Vanilla HB-MCTS*/}
            \STATE \guarantees{$b', \delta_{\bar{D}}^{\beta} \xleftarrow{}$
            \textsc{PruningWithGuarantees}$(b, a, \epsilon_{\bar{D}})$
            /*DA-MCTS. Eq. \eqref{eq:determined_loss}*/}\label{line:apriori}
            \STATE $r \xleftarrow{}$ \textsc{Reward}$(b, a)$
            \STATE $C(ba)\cup \{(b',r)\}$
            \STATE $R \xleftarrow{} r +$\textsc{Rollout}$(b', d-1)$
            \ELSE{}
            \STATE $b',r \xleftarrow{}$ {Sample uniformly from $C(ba)$} 
            \STATE $R$, \guarantees{$\hat{\epsilon}_{\bar{D}}^{hs}$}$
            \xleftarrow{} r +$\textsc{Simulate}$(b', d-1, \epsilon_{\bar{D}})$
            \ENDIF
            \STATE $N(b) \xleftarrow{} N(b) + 1$
            \STATE $N(ba) \xleftarrow{} N(ba) + 1$
            \STATE $Q(ba) \xleftarrow{} Q(ba) + \frac{R-Q(ba)}{N(ba)}$
            \STATE \guarantees{$\hat{\epsilon}_{\bar{D}}^{hs} \xleftarrow{}$
            \textsc{GetGuarantees}$(\hat{\epsilon}_{\bar{D}}^{hs}, \hat{\delta}_{\bar{D}}^{\beta})$
            /*DA-MCTS. Eq. \eqref{eq:estimator_preview}*/} \label{line:posteriori}
            \STATE \textbf{return} $R$, \guarantees{$\hat{\epsilon}_{\bar{D}}^{hs}$}
        \end{algorithmic}
    }
    \end{algorithm}

%% file: 04-Analysis.tex
In this section, we mathematically analyze the impact of pruning on the
performance of the agent. We establish a novel relationship between the complete
and pruned value functions for state-dependent reward functions and provide
bounds on the loss of approximation. Due to restricted space we defer most
proofs and derivations to the supplementary file
\cite{Barenboim23ral_supplementary_b}.

We define $D_t=\{\beta_t^1, \beta_t^2, ..., \beta_t^{|D_t|}\}$ the set of associations at time step
$t$, and $\overline{D}_t\subseteq D_t$ as the subset of hypotheses survived after
the pruning procedure. We define the pruned belief as,
\begin{equation}
    \overline{b}_{t}\triangleq\bar{\mathbb{P}}(x_t\mid H_t) = \sum_{\beta_{t}\in \overline{D}_t}\mathbb{P}(x_t \mid \beta_t ,H_t)\bar{\mathbb{P}}(\beta_{t} \mid H_t),
\end{equation}
where the $\bar{\square}$ notation indicates a pruned distribution after
normalization. This can be explicitly written as,
\begin{equation}
    \overline{b}_{t} = \!\! \!\!\int \limits_{x_{t-1}} \!\!\! \overline{b}_{t-1}\frac{\sum _{\beta _{t} \in \overline{D}_{t}}\mathbb{P}( z_{t} \mid x_{t} ,\beta _{t})\mathbb{P}( \beta _{t} \mid x_{t})\mathbb{P}( x_{t} \mid x_{t-1} , \pi_{t-1})}{\overline{\mathbb{P}}\left( z_{t} \mid H_{t}^{-}\right)},
\end{equation}
where, $\overline{\mathbb{P}}\left( z_{t} \mid H_{t}^{-}\right) =\int
_{x_{t-1:t}}\sum _{\beta _{t} \in \overline{D}_{t}}\mathbb{P}( z_{t} \mid x_{t} ,\beta
_{t})\mathbb{P}( \beta _{t} \mid x_{t})\mathbb{P}( x_{t} \mid x_{t-1} ,\pi (
z_{t-1}))\overline{b}_{t-1}$. Note that the summation is over the pruned set of
hypotheses.
\begin{theorem}\label{theorem:pruning_guarantees} Let time-step $0$ denote the
root of the planning tree. Then, the expected reward for the pruned POMDP,
$\overline{M}$, is bounded with respect to the full POMDP, $M$, through the
factor of the pruned weight values, and the maximum immediate reward,
    \begin{align}
        &\Bigl|\mathbb{E}[ r( b_{t}, a_t)]\! -\!\overline{\mathbb{E}}[ r(\overline{b}_{t}, a_t)]\Bigr| \!\leq \!
        \mathcal{R}_{max}\left[ \delta_0^{\beta} \!+ \!\!\!\sum _{\tau =1}^{t-1}\overline{\mathbb{E}}_{z_{1:\tau}}\!\!\left[ \delta ^{\beta }_{\tau}\right]\right],
    \end{align}
    where $\delta ^{\beta }_{\tau}\triangleq \sum _{\beta_{\tau}\in D_{\tau} \backslash
    \overline{D}_{\tau}}\overline{\mathbb{P}}( \beta _{\tau} \mid H_{\tau})$, i.e. the sum
    of \textit{pruned} hypotheses weights at time-step $\tau$.
\end{theorem}
Crucially, in order to calculate the value of $\delta _{\tau}^{\beta }$, the values
of the hypotheses weights which are descendent of past pruned hypotheses are not required,
as they cannot be obtained without explicitly calculating all hypotheses. More
formally, $\overline{\mathbb{P}}( \beta _{t} \mid
H_{t})=\frac{\mathbb{P}(z_t\mid \beta_t, H_t^-)\sum_{\beta_{0:t-1}\in
\overline{D}}\mathbb{P}(\beta_t\mid \beta_{0:t-1}, H_{t-1}) \mathbb{P}(\beta_{0:t-1}\mid H_t^-)}{\overline{\mathbb{P}}(z_t\mid
H_t^-)}$ has summation only over the survived hypotheses. 

\begin{figure}
	\begin{subfigure}{0.48\textwidth}
		\centering
		\includegraphics[scale=1]{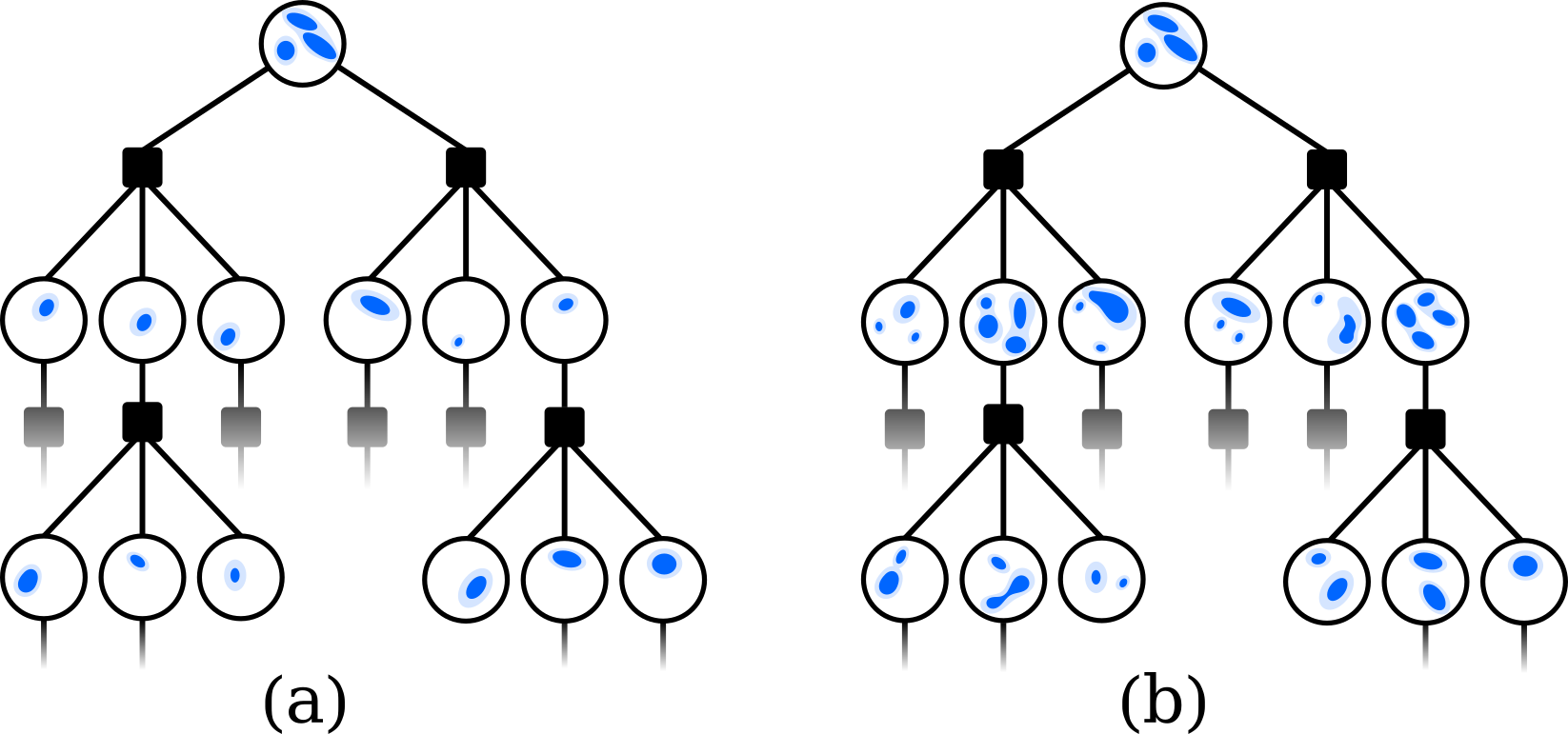}
	\end{subfigure}
	\caption{\scriptsize Planning trees with nodes representing beliefs, and inner blue
	shapes illustrate distributions of the conditional posteriors. (a) A belief
	tree with standard Monte-Carlo estimator leads to an overconfident, fully
	observed data association after a single step. (b) A planning tree with
	Self-Normalized Importance Sampling estimators to account for different
	hypotheses at posterior nodes. \vspace{-12pt}}
	\label{fig:IS_estimator}
\end{figure} 
The generalization of theorem \ref{theorem:pruning_guarantees} to the entire
value function, is straightforward due to linearity of the expectation,
\begin{corollary} \label{cor:theoretical_value_diff}
    Without loss of generality, assume that the time step at the root node of
    the planning tree is $t=0$. Then, for any policy $\pi$, the following holds,
    \begin{equation} \label{cor:theoretical_value_bound}
        \left|V^\pi(b_0)\! -\!\bar{V}^\pi(\bar{b}_0) \right|\! \leq \!
        \mathcal{R}_{max}\!\left[T \delta ^{\beta }_{ 0 }\! + \!\sum _{k=1}^{T}\sum _{\tau =1}^{k}\overline{\mathbb{E}}_{z_{1:\tau }}\!\left[ \delta ^{\beta }_{ \tau}\right]\right].
    \end{equation}
\end{corollary}
For conciseness, we denote this bound as $\epsilon_{\bar{D}}^{hs}$.
As we will derive in the following sections, an equivalent bound can
be derived for estimated value functions, that is,
\begin{equation} \label{eq:estimator_preview}
    |\hat{V}^{{\pi}}(\hat{b}_0) - \hat{\bar{V}}^{{\pi}}(\hat{\bar{b}}_0)|\leq  
    \mathcal{R}_{max}\!\left[T\hat{\delta} ^{\beta }_{ 0}\! + \!\sum _{k=1}^{T}\sum _{\tau =1}^{k}\hat{\overline{\mathbb{E}}}_{z_{1:\tau }}\!\left[ \hat{\delta} ^{\beta }_{ \tau }\right]\right],
\end{equation}
where $\hat{\square}$ denotes an estimator. Similarly, we denote
$\hat{\epsilon}_{\bar{D}}^{hs}$ as the (deterministic) bound for the estimated
value functions.

\subsection{Adaptive Pruning with Performance Guarantees}
The theoretical value bound in Equation \eqref{cor:theoretical_value_bound} and
the estimator value bound in Equation \eqref{eq:estimator_preview} can be used
to provide guarantees for various pruning heuristics, including those presented
in prior work such as \cite{Pathak18ijrr, Barenboim23ral_a} \brown{by providing}
guarantees after the planning session has ended.

In this section, we go a step further, and propose a novel mechanism for
selecting the surviving hypotheses. Unlike previous approaches that use a fixed
budget on the number of allowed hypotheses \cite{Pathak18ijrr}, our algorithm
requires the user to specify the maximum allowable loss, $\epsilon_{\bar{D}}$,
on the value function. Using this allowable loss, our algorithm dynamically
selects the cardinality and instances of hypotheses to prune online, while
maintaining the performance guarantees provided in advance.

To achieve this, we set the value of $\epsilon_{\bar{D}}$ and by construction
determine $\delta^{\beta}_{\tau}$ to be a constant, denoted as $\Delta$, for all
$H_\tau$ and all time steps $\tau$. We use $\Delta$ to determine which
hypotheses to prune in order to meet the budget. The resulting bound can then be
expressed as follows,
\begin{align} \label{eq:determined_loss}
    \left|V^\pi(b_0) -\bar{V}^\pi(\bar{b}_0) \right|&\leq \mathcal{R}_{max}\Delta\!\left[ T + \sum _{k=1}^{T}\sum _{\tau =1}^{k}1\right] \\ 
    &= \mathcal{R}_{max}\Delta\left[\frac{T^2+3T}{2}\right] \triangleq \epsilon_{\bar{D}}.\notag
\end{align}
The hyperparameter $\epsilon_{\bar{D}}$ controls the maximum allowable loss and
is set a priori, as a result $\Delta$ can easily be derived. During planning, we
sum over $\delta^{\beta}_{\tau}$, until its value is as close as possible to
$\Delta$ without crossing its value. The difference between these two values
allows us to obtain a tighter guarantee in hindsight, $\epsilon^{hs}_{\bar{D}}$,
which satisfies the inequality $\epsilon^{hs}_{\bar{D}} \leq
\epsilon_{\bar{D}}$. A similar claim can be made for the sampling-based bound.
The formal derivation of these estimators is presented in the next section.
\vspace{-5pt}

\subsection{Estimated expected reward}
In this section, we first develop an estimator for the value function,
assuming the availability of a complete set of hypotheses at each posterior
belief. Then, we derive a similar, pruning-based estimator. In the next section,
we will show a deterministic relation between the estimators. However, before delving into
the details, we first give a motivation for deriving guarantees with respect to
the estimators.

As stated in Corollary \ref{cor:theoretical_value_diff}, the value function
based on the complete set of hypotheses should not deviate significantly from
the value function based on the pruned hypotheses set, as long as the pruned
hypotheses have low weight values. However, in practice, current
state-of-the-art algorithms cannot compute the full nor the pruned value
functions due to intractable integrals involved with expectations. Online POMDP
algorithms provide performance guarantees based on estimated value functions,
where a sampled set of observations and states approximate expectations and the
belief distribution, e.g., \cite{Silver10nips, Lim22arxiv}.

For clarity, we derive the estimator by considering separately each expected
reward along the planning horizon. Using linearity of the expectation, the
value function may be written as,
\begin{equation}
    V^{\pi}( b_{0})
    =\rho( b_{0} ,\pi_{0}) +\sum _{t=1}^{T}\mathbb{E}_{z_{1:t}}[ \rho ( b_{t}, \pi_t)]. 
\end{equation}
We handle each term in the summation individually, and make the following
proposition as a first step towards deriving an estimated expected reward,
\begin{proposition}  \label{prop:expected_reward_theoretical} Let $z_{1:t}$
    denote an observation sequence, $\rho(b_t,\pi_t)$ be the reward value for a
    given belief, $b_t$ and policy $\pi_t$. The expected reward value can be
    written as,
    \begin{align} \label{eq:expected_reward_theoretical}
        &\mathbb{E}_{z_{1:t}}[ \rho ( b_{t}, \pi_t)] =\\
        &\int _{z_{1:t}}\mathbb{E}_{\beta _{0}}\prod _{\tau =1}^{t}\mathbb{E}_{\beta _{\tau } \mid \beta _{0:\tau -1}}\left[\mathbb{P}\left( z_{\tau } \mid \beta _{0:\tau } ,H_{\tau }^{-}\right) r\left( b_{t}^{\beta } ,\pi _{t}\right)\right],\notag
    \end{align}    
    where $r\left( b_{t}^{\beta } ,\pi _{t}\right)$ denotes the reward value of
    a single hypothesis realization, $\beta_{0:t}$, as shown in equation \eqref{eq:reward_beta}.
\end{proposition}
From the proposition we derive a standard Monte-Carlo sampling approach, where
we iteratively sample sequences of hypotheses $\beta_{0:t}$ and observation
samples, $z_{1:t}$,
\begin{equation}\label{eq:MC}
    \hat{\mathbb{E}}^{MC}_{z_{1:t}}[ \rho ( \hat{b}_{t}, \pi_t)] = \frac{1}{N}\sum_{i}\hat{r}\left( b_{t}^{\beta^i } ,\pi _{t}\right),
\end{equation}
where $\square^{MC}$ denotes Monte-Carlo estimation and $b_{t}^{\beta^i
}\triangleq \mathbb{P}\left(x_t\mid \beta_{0:t}^i, z_{1:t}^i,
\pi_{0:t-1}\right)$. However, since the observation space is continuous,
different realizations of $\beta_{0:t}$, denoted $\beta_{0:t}^i$, will never
sample the same observation sequence $z_{1:t}^i$ twice. In the planning tree, it
means that after an observation sample, there is only a single hypothesis in any
posterior node, resulting in a fully observed data association. However, if the
agent obtains an observation in the real world, the data association ambiguity
is generally not fully resolved. A result, the Monte Carlo sampling
approach is an over-optimistic, erroneous planner which only considers ambiguity
at the root node of the planning tree. See figure \ref{fig:IS_estimator} for an
illustration.

Inspired by \cite{Sunberg18icaps} for standard POMDPs, and
\cite{Barenboim23ral_a} for hybrid POMDPs, we derive an Importance Sampling
(IS) estimator, which may sample observations from different distributions, and
weigh each hypothesis with an importance weight, $\omega \left( z_{\tau
}\right)$. The importance weight reflects the probability of observing $z_t$
given hypothesis $\beta_{0:\tau}$ and history $H_\tau^-$, normalized to the
actual sampling distribution being used, $Q(\cdot)$. We may write equation
\eqref{eq:expected_reward_theoretical} to reflect the change,
\begin{align} \label{eq:theoretical_future_reward}
    &\mathbb{E}_{z_{1:t}}[ \rho ( b_{t}, \pi_t)] =\\
    &\int _{z_{1:t}}\mathbb{E}_{\beta _{0}} \prod _{\tau =1}^{t} Q\left( z_{\tau} \mid H_{\tau}^{-}\right)\mathbb{E}_{\beta _{\tau } \mid \beta _{0:\tau -1}}\left[\omega \left( z_{\tau }\right) r\left( b_{t}^{\beta } ,\pi _{t}\right)\right] \notag
\end{align}
where $\omega \left( z_{\tau }\right) = \frac{\mathbb{P}\left( z_{\tau } \mid
\beta _{0:\tau } ,H_{\tau }^{-}\right)}{Q\left( z_{\tau} \mid
H_{\tau}^{-}\right)}$ and $Q(.)$ is the proposal distribution from which the
sampling-based estimator will sample observations. Clearly, the two terms are
equivalent. 
From \eqref{eq:theoretical_future_reward} we can directly derive the IS-estimator,
\begin{align} \label{eq:IS_estimator}
    &\hat{\mathbb{E}}^{IS}_{z_{1:t}}[ \rho (\hat{b}_{t})] = \hat{\mathbb{E}}_{z_{1:t}}\mathbb{E}_{\beta_{1:t}}[ \hat{r}\left( b_{t}^{\beta }, \pi_t\right)]\triangleq\\ 
    &\sum _{z_{1:t}^{c}}\!\sum _{\beta _{0:t} \in D_{0:t}}\!\!\!\!\mathbb{P}( \beta _{0})\prod _{\tau =1}^{t}\mathbb{P}\left( \beta _{\tau } \mid \beta _{0:\tau -1} ,H_{\tau }^{-}\right)\frac{\omega \left( z_{\tau }^{c}\right)}{N}\hat{r}\left( b_{t}^{\beta }, \pi_t\right)\!,\notag
\end{align}
where, $\hat{r}( b_{t}^{\beta }, \pi_t)$ is the sample-based mean for the
state-reward over the conditional belief, as defined in equation
\eqref{eq:reward_beta}. In contrast to the standard Monte-Carlo estimator
\eqref{eq:MC}, using an importance sampling estimator enables us to reason about
all hypotheses for every observation sequence, shown by the summation over
$\beta_{0:t}$ for each sampled $z_{1:t}^c$.

Although the IS estimator is theoretically justified as a consistent and
unbiased estimator, we make another step in deriving the estimator and use a
Self-Normalized Importance Sampling (SN) estimator, 
\begin{align} \label{def:SN_estimator}
    &\hat{\mathbb{E}}^{SN}_{z_{1:t}}[ \rho (\hat{b}_{t})] = \hat{\mathbb{E}}_{z_{1:t}}\mathbb{E}_{\beta_{1:t}}[ \hat{r}\left( b_{t}^{\beta }, \pi_t\right)]\triangleq\\ 
    &\sum _{z_{1:t}^{c}}\!\!\sum _{\beta _{0:t} \in D_{0:t}}\!\!\!\!\!\mathbb{P}( \beta _{0})\!\!\prod _{\tau =1}^{t}\!\prob{\beta _{\tau } | \beta _{0:\tau -1} ,H_{\tau }^{-}}\frac{\omega \left( z_{\tau }^{c}\right)}{\sum _{z_{\tau }^{k}} \omega \left( z_{\tau }^{k}\right)}\hat{r}\left(\! b_{t}^{\beta }, \pi_t\!\right)\notag
\end{align}
The SN-estimator is no longer unbiased, but is known to be consistent
\cite{Doucet00}. The main reason for that step is to achieve a bounded
deterministic difference between the full and pruned estimators, as we will
describe in the following section. 

Last, we derive a similar estimator for the
\emph{pruned} posterior belief, 
\begin{align} \label{def:pruned_SN_estimator}
    &\hat{\overline{\mathbb{E}}}_{z_{1:t}}\left[ \rho \left(\hat{\overline{b}}_{t}, \pi_t\right)\right] = \hat{\mathbb{E}}_{z_{1:t}}\bar{\mathbb{E}}_{\beta_{1:t}}[ \hat{r}\left( b_{t}^{\beta }, \pi_t\right)]\triangleq\\ 
    &\sum _{z_{1:t}^{c}}\!\!\sum _{\beta _{0:t} \in \overline{D}_{0:t}}\!\!\!\!\!\mathbb{P}( \beta _{0})\!\!\prod _{\tau =1}^{t}\!\prob{ \beta _{\tau } \mid \beta _{0:\tau -1} ,H_{\tau }^{-}}\frac{\omega \left( z_{\tau }^{c}\right)}{\sum _{z_{\tau }^{k}} \omega \left( z_{\tau }^{k}\right)}\hat{r}\left(\! b_{t}^{\beta }, \pi_t\!\right)\!.\notag
\end{align}

\subsection{Estimators analysis}
In this section, we derive a bounded relationship between the full and pruned
estimators. Finally, we discuss how these estimators relate to the theoretical
value function. 
\begin{theorem}\label{theorem:est_to_est_guarantees} Let $\pi$ be a policy,
    then the expected reward for the estimated pruned POMDP,
    $\hat{\overline{M}}$, is bounded with respect to the estimated full POMDP,
    $\hat{M}$, as follows,
    \begin{equation}
        \left| \hat{\mathbb{E}}^{\pi}_{z_{1:t}}[ \rho (\hat{b}_{t})] -\hat{\overline{\mathbb{E}}}^{\pi}_{z_{1:t}}\left[ \rho \left(\hat{\overline{b}}_{t}\right)\right]\right| \!\leq\! \mathcal{R}_{max}\left[\hat{\delta }_{0}^\beta \!+ \!\sum _{\tau =1}^{t}\hat{\delta }_{\tau }^\beta \right]\!.
    \end{equation}
    where, $\hat{\delta }_{\tau
    }^\beta=\hat{\mathbb{E}}_{z_{1:t}^{c}}\overline{\mathbb{E}}_{\beta _{0:t-1}}\sum
    _{\beta _{t} \in D_{t} \backslash \overline{D}_{t}}\mathbb{P}\left( \beta
    _{t} \mid \beta _{0:t-1} ,H_{t}^{-}\right)$ for all $\tau\in[1,t]$
    represents the expected sum of conditional hypotheses' weights which are
    myopically pruned and $\hat{\delta }_{0}^\beta=\sum _{\beta _{0} \in D_{0}
    \backslash \overline{D}_{0}}\mathbb{P}\left( \beta _{0} \mid
    H_{t}^{-}\right)$. 
\end{theorem}
In accordance with the theoretical case, as described in Equation
\eqref{eq:theoretical_future_reward}, to evaluate $\hat{\delta }_{\tau }^\beta$, only
the surviving hypotheses from past time steps are needed. The theorem can be
generalized to the full value function by re-introducing the summation. Under
the assumptions of theorem \ref{theorem:est_to_est_guarantees} the following
holds,
\begin{corollary} \label{cor:bounded_estimated} The difference between the
    estimated value function of the full POMDP, $\hat{M}$, and the estimated
    value function of the pruned POMDP, $\hat{\overline{M}}$, is bounded by,
    \begin{equation}
        |\hat{V}^{{\pi}}(\hat{b}_0) - \hat{\bar{V}}^{{\pi}}(\hat{\bar{b}}_0)|\leq  
        \mathcal{R}_{max}\left[T\hat{\delta} ^{\beta }_{ 0} + \sum _{k=1}^{T}\sum _{\tau =1}^{k} \hat{\delta} ^{\beta }_{ \tau }\right].
    \end{equation}
\end{corollary}
The corollary relates the complete but computationally expensive value function
estimator to the efficient, pruning-based estimator. Both estimators utilize the
same sampled observations since they share the same proposal distribution.

\begin{figure*}[t] 
    \centering
	\begin{subfigure}{0.48\textwidth}
		\includegraphics[width=\textwidth]{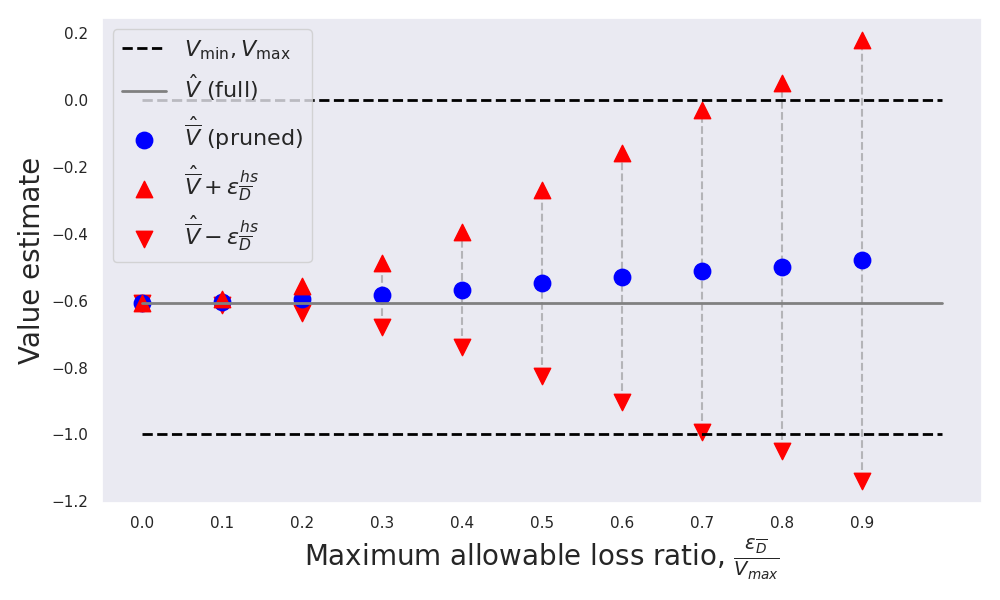}
        \caption{}
        \label{fig:bounds}
	\end{subfigure}
	\begin{subfigure}{0.48\textwidth}
		\includegraphics[width=\textwidth]{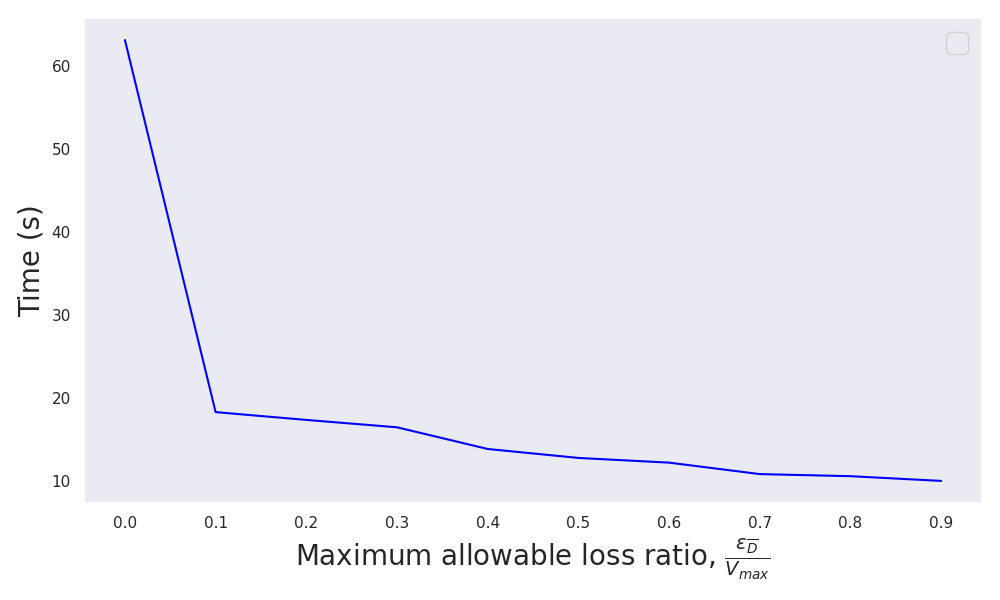}
        \caption{}
        \label{fig:time}
	\end{subfigure} 
    \caption{\scriptsize (a) Bounds of our approach with respect to level of
    simplification. $\hat{V}, \hat{\bar{V}}$ are the value functions of the full
    and pruned estimators respectively. $\hat{\bar{V}} +
    \hat{\epsilon}_{\bar{D}}^{hs}$ represent the bounds of the pruned estimator.
    \brown{$V_{\min}, \backslash V_{\max}$ represent the minimum and maximum
    theoretical values of the value function. All values are normalized with
    respect to $\max\{\left|V_{min}\right|, \left|V_{max}\right|\}$. Here
    $\left|V_{max}\right|\equiv0$ since the reward is defined as the negative Euclidean  distance to goal.} (b) Time for task completion with respect to level of
    simplification. Each level corresponds to the bounds presented in figure
    (a). \vspace{-12pt}}
    \label{fig:timing_bounds}
\end{figure*}

Finding a finite sample algorithm with practical guarantees between the
estimated value function and the theoretical remains an open challenge in the
POMDP literature and is aside from our current contribution. Nevertheless, to
fully justify our approach, we formally state that given such an algorithm,
denoted $\mathcal{A}$, that utilizes the importance sampling estimator defined
in equation \eqref{def:SN_estimator}, our simplified estimator provides a
relationship to the theoretical value function while being more efficient,
\begin{corollary} \label{col:bounded_policy} Let $\pi$ be a policy and let
    $\mathcal{A}$ be a sampling-based estimator for the value function such that
    $|V^{\pi} (b_0) - \hat{V}^{\pi}(\hat{b}_0)| \leq \epsilon_\mathcal{A}$ with
    probability at least $1-\delta_\mathcal{A}$. Then, the loss in the value
    function for the pruned hypotheses is bounded,
    \begin{align}
        &|V^{\pi}(b_0) - \hat{\bar{V}}^{{\pi}}(\hat{\bar{b}}_0)| \leq \\
        &|V^{\pi}(b_0) - \hat{V}^{{\pi}}(\hat{b}_0)| + 
        |\hat{V}^{{\pi}}(\hat{b}_0) - \hat{\bar{V}}^{{\pi}}(\hat{\bar{b}}_0)|
        \leq \epsilon_\mathcal{A} \!+ \hat{\epsilon}^{hs}_{\bar{D}},
    \end{align}
    and holds with probability $1-\delta_\mathcal{A}$. We use $\hat{\epsilon}^{hs}_{\bar{D}}$ as a
    shorthand for the bounds provided in corollary
    \ref{cor:bounded_estimated}.
\end{corollary}

The results established so far hold for any policy, assuming that both the
theoretical and estimated value functions are based on the same policy. However,
planning based on the pruned belief may result in a different policy from the
optimal one for the underlying POMDP. Nevertheless, we demonstrate that the
optimal policy for the pruned and potentially sampled-based POMDP, denoted
$\bar{\pi}$, incurs bounded loss in performance compared to the optimal policy
for the full theoretical POMDP, denoted $\pi^{\star}$.
\begin{corollary} \label{cor:bounded_optimal} Let $\bar{\pi}$ be the optimal
policy for the pruned, possibly sampled-based POMDP and $\pi^{\star}$ be the
optimal policy for the full theoretical POMDP. Then,
    \begin{equation}
        \left|V^{\pi^{\star}}(b_t) - \hat{\bar{V}}^{\bar{\pi}}(\hat{\bar{b}}_t)\right| \leq \brown{2}\left(\epsilon_\mathcal{A} + \hat{\epsilon}^{hs}_{\bar{D}}\right).
    \end{equation}
\end{corollary}
This is an unsurprising result, since the best policy for the pruned
approximation, $\bar{\pi}$, should perform no worse than the optimal policy,
$\pi^\star$, for the simplified POMDP or otherwise it would have been
selected.

%% file: 05-Experiments.tex
\brown{In this section we experiment with different pruning approaches to
validate our findings. We use MCTS as a baseline algorithm and compare multiple
hypothesis pruning approaches to our adaptive scheme.} The experimental
evaluation of our approach consists of two main parts. In the first part, we
validate the proposed bounds and investigate their sensitivity to the level of
simplification chosen. In the second part, we conduct a simulation study to
demonstrate the practical performance gains of our \brown{adaptive pruning}
approach.

\brown{Importantly, we emphasize that the theoretical guarantees presented in
section \ref{sec:analysis} are suitable for other hypotheses-based algorithms as
well, such as \cite{Pathak18ijrr,Barenboim23ral_a}  or PFT-DPW
\cite{Sunberg18icaps} if the latter is adapted to multiple hypotheses.}

\begin{figure*}[h] 
    \centering
	\begin{subfigure}{0.32\textwidth}
		\includegraphics[width=\textwidth]{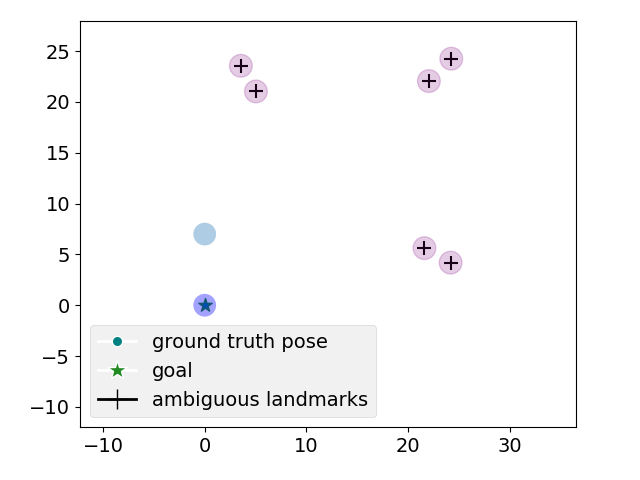}
        \caption{}
        \label{fig:time0}
	\end{subfigure}
	\begin{subfigure}{0.32\textwidth}
		\includegraphics[width=\textwidth]{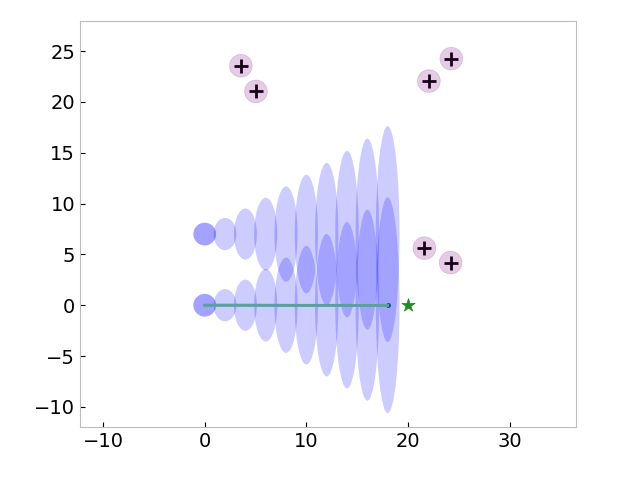}
        \caption{}
        \label{fig:time10}
	\end{subfigure}
	\begin{subfigure}{0.32\textwidth}
		\includegraphics[width=\textwidth]{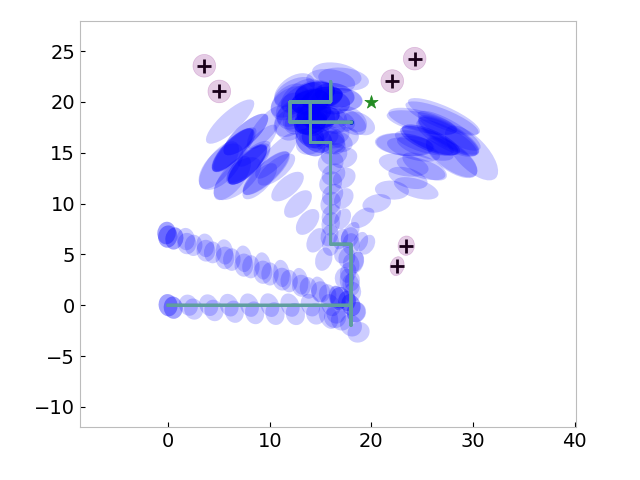}
        \caption{}
        \label{fig:time18}
	\end{subfigure}
    \caption{\scriptsize The figures demonstrate the estimated state of the entire
    trajectory, also known as the smoothing state, of the agent at time $t$
    given the observed history. (a) The prior of the agent given as two Gaussian
    hypotheses. Each Gaussian represented as an ellipse illustrating its
    Covariance, centered around its mean. The landmarks are part of the agent
    state a-priori but has an uncertain location, with ellipses illustrating
    their Covariances. (b) The belief of the agent adjacent to the first
    waypoint before obtaining any observation. (c) The belief of the agent after
    pruning. Non negligible hypotheses differ substantially. \vspace{-12pt}}
    \label{fig:env}
\end{figure*} 

To conduct the simulations, we utilized the GTSAM library \cite{Dellaert12gt} as
our inference engine. Our belief model is based on a Gaussian Mixture Model, in
which each posterior belief in the planning tree corresponds to multiple
instances of GTSAM factor graphs. Each instance represents a conditional
posterior over the continuous part of the belief, $\prob{x_t \mid \beta_{0:t},
H_t}$, while the discrete part of the belief, $\prob{\beta_{0:t}\mid H_t}$, is
maintained as a list of probability values, each corresponds to an hypothesis.
Apart from the pruning method, which is the focus of this section, all
hyper-parameters are shared across all solvers and remain fixed. The planning is
performed in a receding horizon manner, where after each planning session, only
the first action is executed, and all calculations are done from scratch in the
subsequent step.

In the first experiment the belief of the agent included the pose of the agent
and two ambiguous landmarks. The objective of the agent was to reach a target
destination, encoded into the reward function as the expected Euclidean norm
between the agent pose samples and the target. The field of view of the agent was
chosen to be unbounded \brown{and with unlimited sensing range}, that is, at
every time step, the agent obtains an observation from two sources, but cannot
identify its source. In this simple toy example, the number of hypotheses
quickly grows and becomes intractable due to the exponential nature of the
problem. Given a horizon of 10 steps, the number of hypotheses becomes $D_{10} =
2^{10}$, each is a Gaussian conditional distribution. In this and the next
experiments the action space is defined as primitive actions,
up-down-left-right, in a fixed step size.

The estimated value function obtained from the complete set of hypotheses and
the simplified estimator generated using the adaptive pruning approach, as
outlined in Section \ref{sec:analysis}, are illustrated in figure
\ref{fig:timing_bounds}. The solver was endowed with an a-priori budget,
limiting the maximum loss, denoted as $\epsilon_{\overline{D}}$. Based on the
estimator value, the solver determined online which hypotheses to prune and
which to retain.

The results indicate that, as the bounds become looser, i.e., when the value of
$\epsilon_{\overline{D}}$ increases, the computation time efficiency also
increases, \brown{trading off efficiency with performance. As the bounds
increases beyond the value of $0.7$, they become uninformative since the bounds are
larger or smaller than $V_{\max}, V_{\min}$, respectively. On the other hand,}
when the allowable loss budget was set to zero, no hypotheses were pruned,
resulting in identical value estimations for both the pruned and the full
estimators,
\brown{which leads to an identical result as the baseline method of no pruning.}

In the second experiment, we aimed to compare the ability of \brown{different
pruning schemes} to complete the task under a limited time-budget of 20 seconds,
\brown{identical to all solvers}. Specifically, we compare the performance of
our approach to \brown{three types of pruning baselines; no pruning
(Full-HB-MCTS), maintaining a fixed number of hypotheses (K-HB-MCTS) and pruning
below a threshold value (P$_{thresh}$-HB-MCTS). Notably, P$_{thresh}$-HB-MCTS
can be seen as an extension of DA-BSP \cite{Pathak18ijrr}, to an MCTS-based
algorithm instead of Sparse Sampling, as the earlier is known to perform
empirically better. For each pruning method we have experimented with multiple
hyperparameters, P$_{thresh}\in\{0.01, 0.1, 0.3\}$ for P$_{thresh}$-HB-MCTS,
K$\in\{1, 3, 10\}$ for K-HB-MCTS, and $\frac{\epsilon_{\overline{D}}}{V_{\max}}\in
\{0.1, 0.2, 0.5\}$ for DA-MCTS. The best are shown in Table \ref{tab:waypoint}}.

In that experiment, the goal of the agent was to reach an ordered set of
waypoints\brown{, positioned on coordinates $[20,0], [20,20], [0,20]$}, see
figure \ref{fig:env} for an illustration. After performing 60 steps in the
environment, the simulation was restarted. The reward was defined as the
expected sum of distance to the \brown{next waypoint}. The state space was
defined as the agent pose, and the positions of the landmarks. Ambiguous
landmarks were placed in the vicinity of each waypoint to challenge the solvers
by causing an exponential increase in the number of hypotheses.

The results of this experiment are presented in Table \ref{tab:waypoint}. Our
findings indicate that the performance of the HB-MCTS algorithm improved when
the number of hypotheses was reduced. Given the allocated time budget,
maintaining a large set of hypotheses significantly impeded efficiency, leading
to a degradation of the planner's exploration. Conversely, maintaining a single
hypothesis resulted in an overconfident solver that potentially relied on the
wrong association sequence. Our proposed algorithm performed comparably well,
\brown{as it was able} to distinguish between hypotheses with a significant
impact on the value function and those with low impact, which can be pruned.

\begin{table}[htbp]
    \centering
    \caption{\scriptsize Reaching waypoints performance over 10 trials. \brown{The pruning
    hyperparameters chosen for the experiments are $(K=1, $P$_{thresh}=0.1,
    \frac{\epsilon_{\overline{D}}}{V_{\max}}=0.2)$ for K-HB-MCTS,
    \brown{P$_{thresh}$-HB-MCTS} and DA-MCTS respectively.}}
    \label{tab:waypoint}
    \begin{tabular}{l*{4}{S[table-format=2.2,
                                round-mode=places,
                                round-precision=2,
                                table-space-text-post=\,\%]}}
      \toprule
      Algorithm & {Waypoint 1} & {Waypoint 2} & {Waypoint 3} \\
      \midrule
      DA-MCTS (ours) & \textbf{100.0\%} & \textbf{100.0\%} & \textbf{90.0\%} \\
      Full-HB-MCTS & \textbf{100.0\%} & 30.0\% & 20.0\% \\
      K-HB-MCTS & \textbf{100.0\%} & 80.0\% & 60.0\% \\
      \brown{P$_{thresh}$-HB-MCTS} & \textbf{100.0\%} & 80.0\% & 60.0\% \\
      \bottomrule
    \end{tabular}
  \end{table}

%% file: 06-Conclusions.tex
This paper proposes a pruning-based approach for efficient autonomous
decision-making in environments with ambiguous data associations. The approach
models the data association problem as a partially observable Markov decision
process (POMDP) and represents multiple data association hypotheses as a belief
mixture. The challenge of handling the exponential growth in the number of
hypotheses was addressed by pruning the hypotheses while planning, with the
number of hypotheses being adapted based on bounds derived on the value
function.

The results of our evaluations in simulated environments demonstrate the
effectiveness of our approach in handling multi-modal belief hypotheses with
ambiguous data associations. Our method provides a practical solution for
autonomous agents to make decisions in environments with partial observability
and guaranteed performance.

\brown{Future research goals include extending the bounds to hybrid belief
use-cases, improving solver scalability for ambiguous data associations,}
efficient recovery of lost hypotheses, and exploring computational burden
reduction techniques like merging hypotheses with guarantees.

%% file: main.bbl
\begin{thebibliography}{10}
\providecommand{\url}[1]{#1}
\csname url@rmstyle\endcsname
\providecommand{\newblock}{\relax}
\providecommand{\bibinfo}[2]{#2}
\providecommand\BIBentrySTDinterwordspacing{\spaceskip=0pt\relax}
\providecommand\BIBentryALTinterwordstretchfactor{4}
\providecommand\BIBentryALTinterwordspacing{\spaceskip=\fontdimen2\font plus
\BIBentryALTinterwordstretchfactor\fontdimen3\font minus
  \fontdimen4\font\relax}
\providecommand\BIBforeignlanguage[2]{{%
\expandafter\ifx\csname l@#1\endcsname\relax
\typeout{** WARNING: IEEEtran.bst: No hyphenation pattern has been}%
\typeout{** loaded for the language `#1'. Using the pattern for}%
\typeout{** the default language instead.}%
\else
\language=\csname l@#1\endcsname
\fi
#2}}

\bibitem{Pathak18ijrr}
S.~Pathak, A.~Thomas, and V.~Indelman, ``A unified framework for data
  association aware robust belief space planning and perception,'' \emph{Intl.
  J. of Robotics Research}, vol.~32, no. 2-3, pp. 287--315, 2018.

\bibitem{Barenboim23ral_supplementary_b}
\BIBentryALTinterwordspacing
M.~Barenboim, I.~Lev-Yehudi, and V.~Indelman, ``Data association aware pomdp
  planning with hypothesis pruning performance guarantees - supplementary
  material,'' Technion - Israel Institute of Technology, Tech. Rep. [Online].
  Available: \url{https://tinyurl.com/2fekv2pu}
\BIBentrySTDinterwordspacing

\bibitem{Cadena16tro}
C.~Cadena, L.~Carlone, H.~Carrillo, Y.~Latif, D.~Scaramuzza, J.~Neira, I.~D.
  Reid, and J.~J. Leonard, ``Simultaneous localization and mapping: Present,
  future, and the robust-perception age,'' \emph{{IEEE} Trans. Robotics},
  vol.~32, no.~6, pp. 1309 -- 1332, 2016.

\bibitem{Fourie16iros}
D.~Fourie, J.~Leonard, and M.~Kaess, ``A nonparametric belief solution to the
  bayes tree,'' in \emph{IEEE/RSJ Intl. Conf. on Intelligent Robots and Systems
  (IROS)}, 2016.

\bibitem{Tchuiev19iros}
V.~Tchuiev, Y.~Feldman, and V.~Indelman, ``Data association aware semantic
  mapping and localization via a viewpoint-dependent classifier model,'' in
  \emph{IEEE/RSJ Intl. Conf. on Intelligent Robots and Systems (IROS)}, 2019.

\bibitem{Doherty19icra}
K.~Doherty, D.~Fourie, and J.~Leonard, ``Multimodal semantic slam with
  probabilistic data association,'' in \emph{2019 international conference on
  robotics and automation (ICRA)}.\hskip 1em plus 0.5em minus 0.4em\relax IEEE,
  2019, pp. 2419--2425.

\bibitem{Sunberg18icaps}
Z.~Sunberg and M.~Kochenderfer, ``Online algorithms for pomdps with continuous
  state, action, and observation spaces,'' in \emph{Proceedings of the
  International Conference on Automated Planning and Scheduling}, vol.~28,
  no.~1, 2018.

\bibitem{Somani13nips}
A.~Somani, N.~Ye, D.~Hsu, and W.~S. Lee, ``Despot: Online pomdp planning with
  regularization.'' in \emph{NIPS}, vol.~13, 2013, pp. 1772--1780.

\bibitem{Shienman22icra}
M.~Shienman and V.~Indelman, ``D2a-bsp: Distilled data association belief space
  planning with performance guarantees under budget constraints,'' in
  \emph{IEEE Intl. Conf. on Robotics and Automation (ICRA)}, 2022.

\bibitem{Shienman22isrr}
------, ``Nonmyopic distilled data association belief space planning under
  budget constraints,'' in \emph{Proc. of the Intl. Symp. of Robotics Research
  (ISRR)}, 2022.

\bibitem{Barenboim23ral_a}
M.~Barenboim, M.~Shienman, and V.~Indelman, ``Monte carlo planning in hybrid
  belief pomdps,'' \emph{IEEE Robotics and Automation Letters}, vol.~8, no.~8,
  pp. 4410--4417, 2023.

\bibitem{Doucet00}
A.~Doucet, N.~{de Freitas}, and N.~Gordon, Eds., \emph{Sequential {M}onte
  {C}arlo Methods In Practice}.\hskip 1em plus 0.5em minus 0.4em\relax New
  York: Springer-Verlag, 2001.

\bibitem{Kocsis06ecml}
L.~Kocsis and C.~Szepesv{\'a}ri, ``Bandit based monte-carlo planning,'' in
  \emph{European conference on machine learning}.\hskip 1em plus 0.5em minus
  0.4em\relax Springer, 2006, pp. 282--293.

\bibitem{Silver10nips}
D.~Silver and J.~Veness, ``Monte-carlo planning in large pomdps,'' in
  \emph{Advances in Neural Information Processing Systems (NIPS)}, 2010, pp.
  2164--2172.

\bibitem{Lim22arxiv}
M.~H. Lim, T.~J. Becker, M.~J. Kochenderfer, C.~J. Tomlin, and Z.~N. Sunberg,
  ``Generalized optimality guarantees for solving continuous observation pomdps
  through particle belief mdp approximation,'' \emph{arXiv preprint
  arXiv:2210.05015}, 2022.

\bibitem{Dellaert12gt}
F.~Dellaert, ``Factor graphs and gtsam: A hands-on introduction,'' Georgia
  Institute of Technology, Tech. Rep., 2012, gTSAM.

\end{thebibliography}
